\documentclass[10pt,twocolumn,letterpaper]{article}

\usepackage{cvpr}              
\usepackage[accsupp]{axessibility}  
\usepackage{lipsum}
\usepackage[dvipsnames]{xcolor}

\definecolor{cvprblue}{rgb}{0.21,0.49,0.74}
\usepackage[pagebackref,breaklinks,colorlinks,citecolor=cvprblue]{hyperref}

\usepackage[dvipsnames]{xcolor}  

\def\paperID{8726} 
\def\confName{CVPR}
\def\confYear{2025}

\title{FreeScene: Mixed Graph Diffusion for 3D Scene Synthesis from Free Prompts}

\author{
    \makebox[\textwidth]{ 
        Tongyuan Bai$^{1}$ \quad Wangyuanfan Bai$^{1}$ \quad  Dong Chen$^{1}$ \quad Tieru Wu$^{1,3}$ \quad Manyi Li$^{2}$ \quad  \quad Rui Ma$^{1,3}$\footnotemark[1] 
    } \\
    \makebox[\textwidth]{{ $^1$School of Artificial Intelligence, Jilin University}} \\
    \makebox[\textwidth]{{ $^2$School of Software, Shandong University}} \\
    \makebox[\textwidth]{{ $^3$Engineering Research Center of Knowledge-Driven Human-Machine Intelligence, MOE, China}} 
}

\usepackage{tcolorbox} 
\usepackage{xcolor}
\usepackage{diagbox}
\usepackage{float}
\usepackage{graphicx}
\usepackage{stfloats}
\usepackage{multicol}
\usepackage{multirow}
\usepackage{tabularx}
\usepackage{booktabs} 
\usepackage{multicol}
\usepackage{cuted}
\begin{document} 

\twocolumn[{%
\renewcommand\twocolumn[1][]{#1}%
\maketitle
\vspace*{-20pt}
\begin{center}
\centering
\includegraphics [width=1\linewidth]{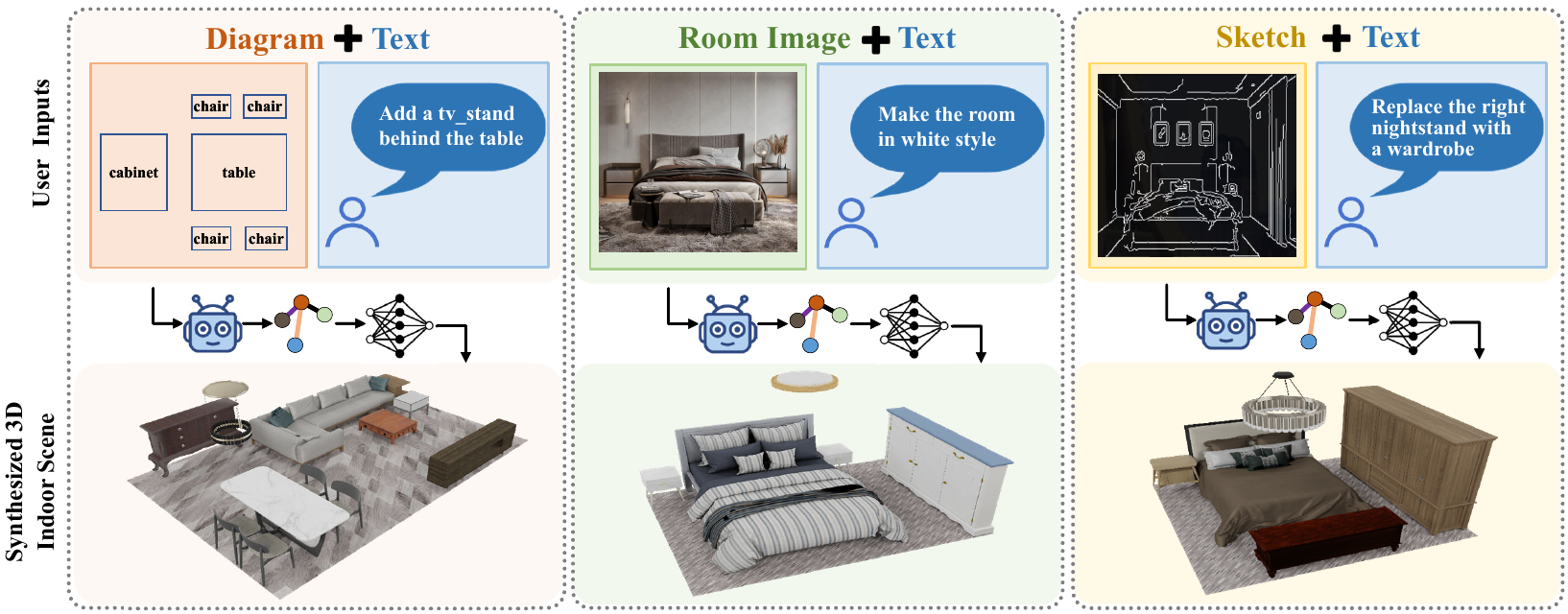}
\captionof{figure}{
    We introduce FreeScene, a user-friendly controllable indoor scene synthesis framework that allows convenient and effective control with free-form user inputs, including text and/or different types of images (top-view diagram in the left, realistic photograph in the middle, sketch in the right). Leveraging a multimodal agent  (\raisebox{-0.2em}{\includegraphics[height=1em]{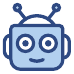}}) to extract partial graph priors (\raisebox{-0.2em}{\includegraphics[height=1em]{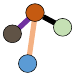}}) from user inputs, FreeScene subsequently generates a complete and reasonable scene that conforms to the graph via  a generative diffusion model (\raisebox{-0.2em}{\includegraphics[height=1em]{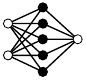}}). 
}
\label{fig:teaser}
\end{center}
}]
\renewcommand{\thefootnote}{\fnsymbol{footnote}}
\footnotetext[1]{Corresponding author}
\renewcommand{\thefootnote}{\arabic{footnote}} 

\begin{abstract}
 
Controllability plays a crucial role in the practical applications of 3D indoor scene synthesis. Existing works either allow rough language-based control, that is convenient but lacks fine-grained scene customization, or employ graph-based control, which offers better controllability but demands considerable knowledge for the cumbersome graph design process. To address these challenges, we present \textbf{FreeScene}, a user-friendly framework that enables both convenient and effective control for indoor scene synthesis. Specifically, FreeScene supports free-form user inputs including text description and/or reference images, allowing users to express versatile design intentions. The user inputs are adequately analyzed and integrated into a graph representation by a VLM-based Graph Designer. We then propose MG-DiT, a Mixed Graph Diffusion Transformer, which performs graph-aware denoising to enhance scene generation. Our MG-DiT not only excels at preserving graph structure but also offers broad applicability to various tasks, including, but not limited to, text-to-scene, graph-to-scene, and rearrangement, all within a single model. Extensive experiments demonstrate that FreeScene provides an efficient and user-friendly solution that unifies text-based and graph-based scene synthesis, outperforming state-of-the-art methods in terms of both generation quality and controllability in a range of applications. Project page: https://cangmushui.github.io/FreeScene-io/
\end{abstract}    
\section{Introduction}
\label{sec:intro}


Indoor scene synthesis has demonstrated immense applications across various domains such as game design, VR/AR, and robotics, where the goal is to produce reasonable and plausible scenes that meet user requirements such as text description, sketches, images, etc. To date, there has been a significant body of related works, utilizing either graphical probabilistic models~\cite{yu2015clutterpalette, xu2002constraint, yu2011make, merrell2011interactive, fisher2012example, kermani2016learning, liu2014creating, zhang2021fast}  or deep generative models~\cite{wang2018deep, ritchie2019fast, zhou2019scenegraphnet, wang2019planit, paschalidou2021atiss, li2019grains, yang2021scene, gao2023scenehgn, zhang2020deep, yang2021indoor} to learn the data distribution of indoor scenes. Recently, as the diffusion model~\cite{ho2020denoising, song2019generative, song2020score} has shown excelling generative performance compared to VAE~\cite{kingma2013auto} and GANs~\cite{goodfellow2014generative}, it is also applied to indoor scene synthesis by iteratively denoising the scene representations, enabling the generation of plausible and high-quality scene results~\cite{tang2024diffuscene, yang2024physcene, hu2024mixed, maillard2024debara}.

However, controllability, as a critical factor for user engagement, has yet to be adequately addressed to provide both convenient and precise control over indoor scene synthesis with diffusion models. The recent DiffuScene~\cite{tang2024diffuscene}  incorporates text control through a cross-attention mechanism on U-Net~\cite{ronneberger2015u}, but only obtains a rough consistency between the text conditions and generated scenes without precise control. InstructScene~\cite{lin2024instructscene} introduces a discrete graph diffusion model to convert text conditions into scene graphs which offer fine-grained control. Although the scene graphs align better with the generated scenes, their text-to-graph requires users to provide a detailed description of the objects and their relations, while still being prone to information loss.

In this paper, we introduce FreeScene, an indoor scene synthesis framework with convenient and effective control. We allow users to express their thoughts freely with a mixture of text descriptions and images, such as top-view diagrams, realistic photos, and sketches. Our framework includes a VLM-based Graph Designer, which conducts stepwise multimodal understanding and reasoning to infer object arrangement as scene graphs, and a mixed graph diffusion transformer model coined as MG-DiT, which refines the graph to generate complete and reasonable scene configurations. Specifically, MG-DiT uses constrained sampling to preserve input conditions throughout the denoising process, enabling a variety of applications such as text-to-scene and graph-to-scene, with a single model. To the best of our knowledge, this is the first approach capable of addressing all these tasks with a unified diffusion model. Extensive experiments demonstrate that FreeScene outperforms state-of-the-art approaches in both generation quality and controllability across many applications.

Our contributions are summarized as follows:
\begin{itemize} 
\item We introduce the FreeScene framework, which utilizes a graph designer capable of interpreting the free-form user inputs as graphs and a diffusion model to generate results for the user-friendly controllable scene synthesis.
\item We propose the MG-DiT model, which employs a mixed graph diffusion process with constrained sampling to enable a variety of applications with a single model.
\item We conduct comparison experiments on a series of applications to demonstrate the flexibility of our approach and the superior performance over state-of-the-art methods.
\end{itemize}

\section{Related Work}
\label{sec:related_work}
\noindent \textbf{Indoor Scene Synthesis.} The goal of indoor scene synthesis~\cite{Patil2024AdvancesID} is to generate diverse and plausible indoor scene layouts. Some early works~\cite{yu2011make, merrell2011interactive, fisher2012example, liu2014creating, kermani2016learning} adopt probabilistic models to parameterize the distribution of indoor scenes and sample novel scenes. However, due to the various object relationships and spatial constraints within indoor scenes, they suffer from the time-consuming optimization process and limited range of generated scenes. 

Many studies have demonstrated the effectiveness of training deep neural networks on different synthetic large-scale indoor scene datasets~\cite{song2017semantic, li2018interiornet, fu20213d}. The architectures of deep generative models have ranged from convolutional networks~\cite{wang2018deep, ritchie2019fast}, recursive neural networks~\cite{li2019grains, gao2023scenehgn}, graph neural networks~\cite{zhou2019scenegraphnet, wang2019planit}, to Transformers~\cite{paschalidou2021atiss} and diffusion models~\cite{Wei2023LEGONetLR, tang2024diffuscene}. Among the recent works, ATISS~\cite{paschalidou2021atiss} represents the indoor scenes as object sets and leverages the permutation equivariance of the transformer to learn the layout distribution. DiffuScene~\cite{tang2024diffuscene} trains a diffusion model to progressively denoise the layout attributes and finally generate plausible indoor scenes. To date, the attention mechanism and denoising diffusion model have been validated to be effective in capturing the complex relative placements between objects within the indoor scenes.

In parallel, some pioneer works~\cite{yang2024holodeck, fu2025anyhome} study to use the pre-trained Large Language Models (LLMs) to facilitate open-vocabulary scene synthesis. They mainly focus on prompting the LLMs to generate complete scene configurations from the given simple and abstract descriptions, and directly parse the responses into scene layouts with heuristic optimizations. However, these works lack fine-grained and precise control over the generated scenes.

\noindent\textbf{Controllable Scene Synthesis.} Controllability is crucial for generating indoor scenes that meet user expectations. Many related works have explored scene synthesis conditioned on text descriptions ~\cite{seversky2006real, chang2014learning, chang2014interactive, chang2015text, chang2017sceneseer, ma2018language}, sketches~\cite{shin2007magic, lee2008sketch, xu2013sketch2scene}, images~\cite{izadinia2017im2cad, huang2018cooperative, Nie_2020_CVPR}, etc. They often exploit structured scene graphs to represent the specified conditions and capture object relation priors to guarantee the semantic alignment between input conditions and synthesized scenes.

Graph-based networks have shown significant potential in conditional scene synthesis. PartIT~\cite{wang2019planit} utilizes a graph convolutional network to synthesize relation graphs, followed by an image-based convolution network to instantiate novel scenes, enhancing both the diversity and plausibility. Some other related works, such as SceneGraphNet~\cite{zhou2019scenegraphnet} and CommonScenes~\cite{zhai2024commonscenes}, directly take scene graphs as conditions and use graph neural networks with message passing modules to complete layout attributes of the graphs. Graph-to-3D~\cite{dhamo2021graph} and EchoScene~\cite{zhai2024echoscene} incorporate both the object shape and their placement while facilitating information exchange between them to synthesize diverse scenes. InstructScene~\cite{lin2024instructscene} quantifies various object features with a codebook and proposes a discrete semantic graph diffusion which significantly improves the controllability and fidelity of the synthesized indoor scenes.

\noindent\textbf{Diffusion Models for Layout Synthesis.} The diffusion model~\cite{ho2020denoising, song2019generative, song2020score} has made remarkable achievement in a variety of content creation applications. As previously mentioned, some works~\cite{Wei2023LEGONetLR, tang2024diffuscene} directly utilize the diffusion model for indoor scene synthesis. Upon the basic design of diffusion models, PhyScene~\cite{yang2024physcene} uses several guidances for the diffusion model to improve the physical plausibility and interactivity. EchoScene~\cite{zhai2024echoscene} associates each object node with a denoising process and enables collaborative information exchange for better controllability and fidelity.

Indoor scene representation plays a crucial role during the iterative denoising with diffusion models. The straightforward solution is to represent scene layouts as a set of object tokens in the continuous attribute vector space~\cite{tang2024diffuscene, yang2024physcene} or latent space~\cite{zhai2024echoscene}. With the development of Discrete Denoising Diffusion Probabilistic Models (D3PM)~\cite{austin2021structured}, several other domains that involve graph structures have started employing discrete~\cite{qin2023sparse, liu2024graph} or mixed diffusion~\cite{vignac2023midi} models to accomplish generative tasks, showcasing their potential in producing structured and high-quality outputs. Some works quantify the object attributes~\cite{Inoue2023layoutdm} or features~\cite{lin2024instructscene} as discrete tokens and utilize discrete diffusion for layout synthesis. Further, MiDiffusion~\cite{hu2024mixed} takes object categories as discrete tokens and placement attributes as continuous values,  and proposes to learn their distribution simultaneously to produce scenes with high fidelity and plausibility.

\section{Preliminary}

In this section, we concisely introduce the diffusion models in the continuous and discrete domains, which form the foundation for the derivation of our loss function.
\label{sec:preliminary}

\noindent\textbf{Continuous Diffusion Model.}
The Denoising Diffusion Probabilistic Model (DDPM)~\cite{ho2020denoising} proposed the derivation for the forward diffusion process and reverse denoising process in continuous space. By fitting the posterior distribution of the reverse denoising with a neural network, we can obtain data samples that conform to the true distribution. DDPM initially injects Gaussian noise into the continuous variable $x_0$ iteratively, such that by time $T$, the data distribution approximates a standard normal distribution:
\begin{small} 
\begin{equation}
    q(\boldsymbol{x}_t|\boldsymbol{x}_{t-1}):=\mathcal{N}(\boldsymbol{x}_t;\sqrt{1-\beta_t}\boldsymbol{x}_{t-1},\beta_t\boldsymbol{I}),
\end{equation}
\end{small}%
where $\beta_{1},\ldots,\beta_{T}\in(0,1)$. The conditional distribution of the forward process at timestep $t$ can be expressed as:
\begin{small} 
\begin{equation}
    q(\boldsymbol{x}_t|\boldsymbol{x}_0)=\mathcal{N}(\boldsymbol{x}_t;\sqrt{\bar{\alpha}_t}\boldsymbol{x}_0,(1-\bar{\alpha}_t)\boldsymbol{I}),
\end{equation}
\end{small}%
where $\alpha_{t}:=1-\beta_{t}\mathrm{~and~}\bar{\alpha}_{t}:=\prod_{s=1}^{t}\alpha_{s}$. 
For the posterior distribution during the reverse denoising process, it is also represented as Gaussian distribution:
\begin{small} 
\begin{equation}
    q(\boldsymbol{x}_{t-1}|\boldsymbol{x}_t,\boldsymbol{x}_0)=\mathcal{N}(\boldsymbol{x}_{t-1};\tilde{\boldsymbol{\mu}}_t(\boldsymbol{x}_t,\boldsymbol{x}_0),\tilde{\beta}_t\boldsymbol{I}),
\end{equation}
\end{small}%
where $\tilde{\boldsymbol{\mu}}_t(\boldsymbol{x}_t,\boldsymbol{x}_0) := \frac{\sqrt{\bar{\alpha}_{t-1}}\beta_t}{1-\bar{\alpha}_t}\boldsymbol{x}_0+\frac{\sqrt{\alpha_t}(1-\bar{\alpha}_{t-1})}{1-\bar{\alpha}_t}\boldsymbol{x}_t$ and the variance is defined as $\tilde{\beta}_t := \frac{1-\bar{\alpha}_{t-1}}{1-\bar{\alpha}_t}\beta_t$.

\noindent\textbf{Discrete Diffusion Model.}
D3PM \cite{austin2021structured} defines a state transition matrix and introduces a [MASK] state to enable the forward diffusion and reverse denoising in the discrete space. For a discrete variable $z$ with $K$ categories, the state transition matrix is a $\mathbb{R}^{(K+2)\times (K+2)}$ matrix, where the extra two dimensions are for the empty and mask states, i.e.,
\begin{small} 
\begin{equation}
    \left.\mathbf{Q}_t:=\left[\begin{array}{ccccc}\alpha_t+\beta_t&\beta_t&\cdots&\beta_t&0\\\beta_t&\alpha_t+\beta_t&\cdots&\beta_t&0\\\vdots&\vdots&\ddots&\beta_t&0\\\beta_t&\beta_t&\beta_t&\alpha_t+\beta_t&0\\\gamma_t&\gamma_t&\gamma_t&\gamma_t&1\end{array}\right.\right].
\end{equation}
\end{small}%
Therefore, at a given timestep $t$, a discrete variable $\boldsymbol{z}_t$ (represented as a one-hot row vector) has a probability $\gamma_t$ of being masked in the subsequent time step, a probability $\alpha_t$ of remaining unchanged, while the remaining probability $(1-\gamma_t-\alpha_t)$ is uniformly transited to the other $K+1$ states (1 for empty). Then the discrete forward diffusion process can be defined as a conditional probability distribution:
\begin{small} 
\begin{equation}
q(\boldsymbol{z}_t|\boldsymbol{z}_{t-1})=\mathrm{Cat}(\boldsymbol{z}_t;\boldsymbol{p}=\boldsymbol{z}_{t-1}\boldsymbol{Q}_t),
\end{equation}
\end{small}%
where $\mathrm{Cat}(\boldsymbol{z};\boldsymbol{p})$ is a categorical distribution over the one-hot row vector $\boldsymbol{z}$ with probabilities for each possible outcome given by the row vector $\boldsymbol{p}$. And the entire forward process can be parameterized as follows:
\begin{small} 
\begin{equation}
q(\boldsymbol{z}_t|\boldsymbol{z}_0)=\mathrm{Cat}\left(\boldsymbol{z}_t;\boldsymbol{p}=\boldsymbol{z}_0\overline{\boldsymbol{Q}}_t\right),
\end{equation}
\end{small}%
where $\overline{Q}_t=Q_1Q_2\ldots Q_t$. Accordingly, the posterior distribution of the denoising process for discrete variables is
\begin{small}
\begin{equation}
\begin{aligned}
&q(\boldsymbol{z}_{t-1}|\boldsymbol{z}_t,\boldsymbol{z}_0)=\frac{q(\boldsymbol{z}_t|\boldsymbol{z}_{t-1},\boldsymbol{z}_0)q(\boldsymbol{z}_{t-1}|\boldsymbol{z}_0)}{q(\boldsymbol{z}_t|\boldsymbol{z}_0)}\\
&=\mathrm{Cat}\left(\boldsymbol{z}_{t-1};\boldsymbol{p}=\frac{\boldsymbol{z}_t\boldsymbol{Q}_t^\top\odot\boldsymbol{z}_0\overline{\boldsymbol{Q}}_{t-1}} 
{\boldsymbol{z}_0\overline{\boldsymbol{Q}}_t\boldsymbol{z}_t^\top}\right).
\end{aligned}
\end{equation}
\end{small}
\section{Method}
\begin{figure*}
\centering
\includegraphics[width=0.95\linewidth]{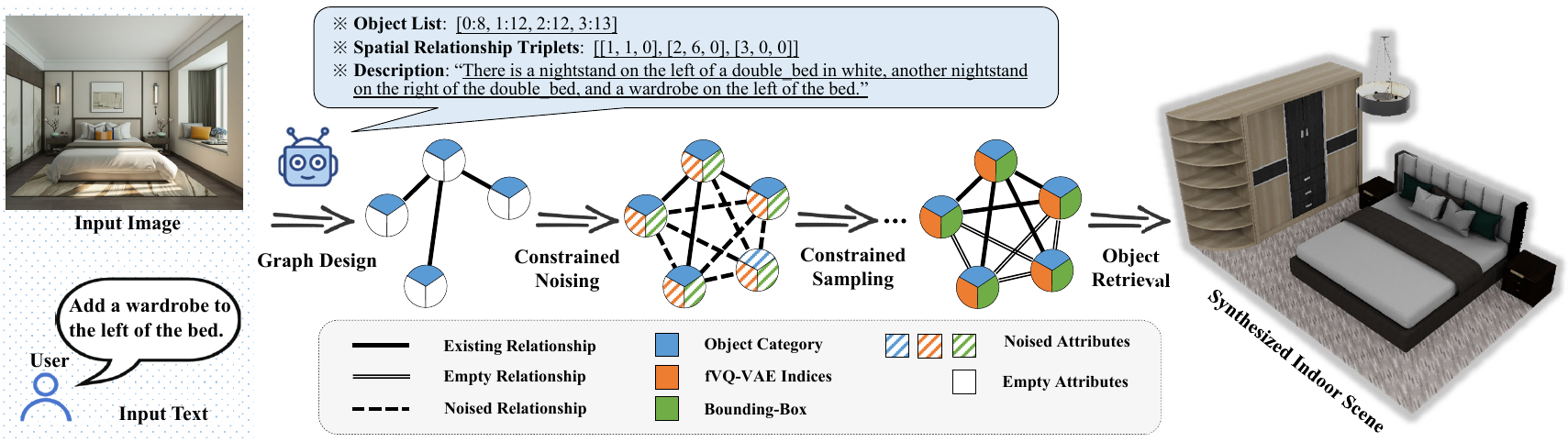}
\caption{
    Overview of FreeScene.
    Given a reference image, text, or either of them, a carefully designed agent called the \textbf{Graph Designer} analyzes the objects and their spatial relationships, constructing a partial graph prior that captures both the object types and their spatial relationships.
    Next, leveraging the proposed \textbf{MG-DiT}, we apply a constrained sampling method that preserves the integrity of the graph prior throughout the noising and denoising processes, while generating the remaining parts.
    Ultimately, we obtain the complete room layout and perform object retrieval from the 3D-FUTURE dataset \cite{fu20213dfu}, retrieving each object with the most similar OpenCLIP \cite{liu2024openshape} feature (derived from the generated fVQ-VAE \cite{lin2024instructscene} indices) within the same category.
}
\label{fig:pipeline}
\vspace*{-14pt}
\end{figure*}

We start with problem formulation (Section \ref{problem_formulation}) of indoor scene synthesis and then introduce our FreeScene framework. As shown in  Figure \ref{fig:pipeline}, the framework comprises two key components: one is a carefully engineered agent named Graph Designer (Section \ref{graph_designer}), which analyzes the given text and/or images, and composites the extracted object and relations into a partial graph; the other is a generative model named MG-DiT (Section \ref{mixed_graph}), which populates the partial graph to generate complete layouts of plausible and coherent  indoor scenes. Finally, we describe various applications derived from MG-DiT based on constrained sampling, without the need for further training (Section \ref{application}).

\subsection{Problem Formulation}
\label{problem_formulation}

We define $\mathcal{S}\!:=\!\{\mathcal{S}_1, \mathcal{S}_2, \cdots, \mathcal{S}_k\}_{k=1}^{K}$ to represent the indoor scene dataset, where each scene $\mathcal{S}_k:=\{\mathcal{O}_k, \mathcal{G}_k\}$ consists of an object collection $\mathcal{O}_k=\{o_i^k\}_{i=1}^N$ with $N$ objects and an edge collection $\mathcal{G}_k\!=\!\{e_{i\rightarrow j}\in\{ 1, 2, \cdots,Q\}|i,j\in\{1,2,\cdots,N\},i\neq j\}$, where $e_{i\rightarrow j}$ denotes the spatial relationship from the $i_{th}$ objects to the $j_{th}$ out of $Q$ discrete labels. For objects, such as $o_{i}^{k}$ denoting the $i_{th}$ object in the $k_{th}$ scene, we have $o_{i}^{k}:=(c_{i}^{k}, v_{i}^{k}, s_{i}^{k}, t_{i}^{k}, r_{i}^{k})$ containing the discrete attributes, i.e. object category $c_{i}^{k}\in\{1,2,\cdots,P\}$ in $P$ discrete labels, feature indices $v_{i}^{k}\in\{1,2,\cdots,K_f\}^{n_f}$ with $n_f$ tokens in $K_f$ discrete labels, and continuous attributes, i.e. object size $s_{i}^{k}\in\mathbb{R}^3$, position $t_{i}^{k}\in\mathbb{R}^3$, and orientation $r_{i}^{k}\in\mathbb{R}^2$ (in the form of $\cos{\theta}$ and $\sin{\theta}$), respectively. Specifically, the feature indices $v_{i}^{k}$ refer to $n_f$ discrete tokens quantized from the features stored in the codebook $\mathcal{Z}\in\mathbb{R}^{K_f\times d_z}$, for which we leverage fVQ-VAE~\cite{lin2024instructscene} to compress the OpenCLIP~\cite{liu2024openshape} features of the objects. Finally, assuming $y_k$ is the textual description of the $k_{th}$ scene in the dataset, our target is to optimize the parameters $\theta$ to maximize the following log-likelihood:
\begin{small}
\begin{equation}
\log p_{\theta}(\mathcal{S}) =\sum_{k=1}^{K}\log p_{\theta}(\mathcal{S}_{k}|y_{k})  
=\sum_{k=1}^{K}\log p_{\theta}(\mathcal{O}_{k},\mathcal{G}_{k}|y_{k}).
\end{equation}
\end{small}

\subsection{Graph Designer}
\label{graph_designer}
\begin{figure}
\centering
\includegraphics[width=0.95\linewidth]{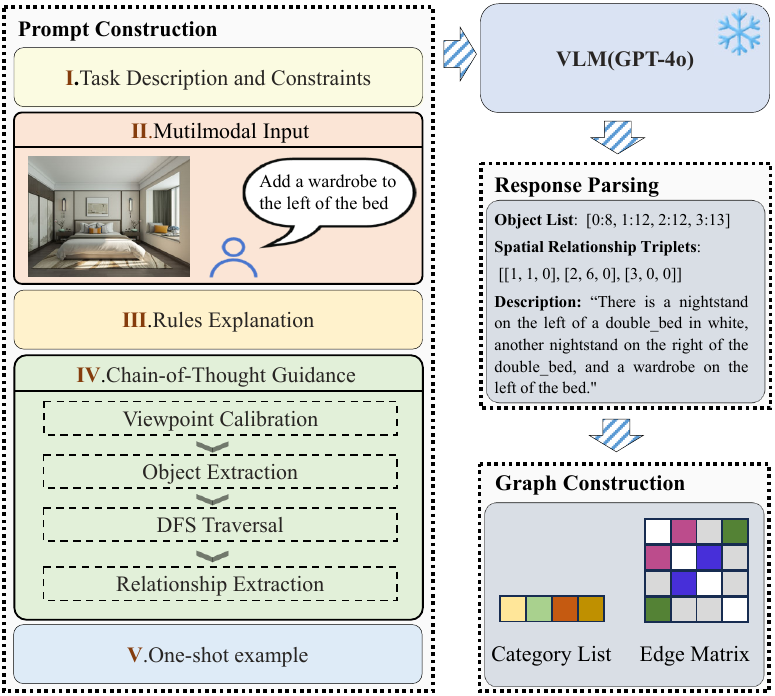}
\caption{Graph Designer. 
The Graph Designer utilizes a one-shot Chain-of-Thought (CoT) method to guarantee the accuracy of the extraction results. It then parses and preprocesses the VLM responses to ensure compatibility with the MG-DiT inputs.
}
\label{fig:graphdesigner}
\vspace{-20pt}
\end{figure}

Our Graph Designer leverages pre-trained Vision-Language Models (VLMs) to extract the objects and their spatial relationships from user-provided text and/or image prompts. The VLM responses are then transformed into the graph representation, which forms the partial graph condition for the subsequent diffusion model for seamless integration and effective processing. 
The partial graph is defined as a list of object categories $\{c_{0},c_{1},\ldots,c_{N^-}\}$ and an edge collection $\mathcal{G}^-$ forming a edge matrix, complemented with a global textual description. Here, $N^-$ represents the number of objects extracted from the free-form user inputs, which acts as a subset of the final scene, and $\mathcal{G}^-$ denotes the relationships corresponding to these extracted objects.

\noindent\textbf{Prompt Construction.} In order to ensure the output data format and efficient information extraction, we develop a one-shot Chain-of-Thought (CoT) prompt template for the VLMs in which a one-shot formated example is given to regulate the output format following \cite{yang2024holodeck}. Moreover, we exploit CoT \cite{wei2022chain} to steer the VLMs to tackle the inference task logically and systematically. Our one-shot CoT prompt consists of the following components: (1) \emph{Task description and constraints}, which outline the task objectives and specify the types of objects and relationships that might be involved. (2) \emph{Multimodal user input} in the form of text and/or image, where the image can be a real or synthesized image, a line drawing sketch, or a top-view diagram. (3) \emph{Rule explanation}, which specifies certain rules governing the response generation process and establishes connections between these rules and the output through detailed descriptions. (4) \emph{Chain-of-Thought guidance}, which provides the step-by-step hint to guide the systematic reasoning process of the VLM. (5) \emph{One-shot example}, a demonstration with a strictly defined format. We instruct the VLMs to adhere to this format, ensuring seamless subsequent parsing.


\noindent\textbf{Chain-of-Thought Guidance.}
To enhance the VLM to stably and precisely infer graph conditions from the free-form user inputs, we formulate the Chain-of-Thought guidance, which consists of the following steps: (1) \emph{Viewpoint calibration}. We instruct the VLM to determine the viewpoint of input images, so that subsequent relationship judgments can accurately distinguish spatial relationships corresponding to different viewpoints. (2) \emph{Object extraction}. We guide the VLM to identify the objects belonging to a set of pre-defined categories from the given image and text. (3) \emph{DFS traversal}. Since calculating pairwise relationships between all objects often leads to highly disorganized outputs or the omission of critical relationships, we design a detailed set of instructions to guide the VLM to start from a key root node and use a depth-first search (DFS) approach to hierarchically traverse all objects, facilitating an organized relationship extraction. (4) \emph{Relationship extraction}. Based on the traversal results from previous step, we individually evaluate each relationship between the parent and child nodes, then integrate the findings to produce the final output.

\noindent\textbf{Graph Construction.}
The VLM output consists of an object list, a relation triplet list, and a textual description of the room. As shown in Figure \ref{fig:graphdesigner}, the object list contains the indices and categories of the objects. The relation triplet list contains the relation type and indices of the two adjacent objects, forming a edge matrix. The outputs are parsable and can be converted to a graph representation, thanks to the one-shot example provided in the prompt. We utilize GPT-4o and apply regular expressions to parse its outputs. The full prompts are provided in Supplementary.


\subsection{Mixed Graph Diffusion Transformer}
\label{mixed_graph}
We propose Mixed Graph Diffusion Transformer (MG-DiT) to facilitate a variety of applications including text-to-scene generation, zero-shot graph-to-scene generation, etc. with a single model through constrained sampling. During the inference phase, we can seamlessly input the combination of the object category list, edge matrix, and textual description produced by the Graph Designer as the partial graph condition of MG-DiT for indoor scene synthesis.
\begin{figure}
\centering
\includegraphics[width=0.95\linewidth]{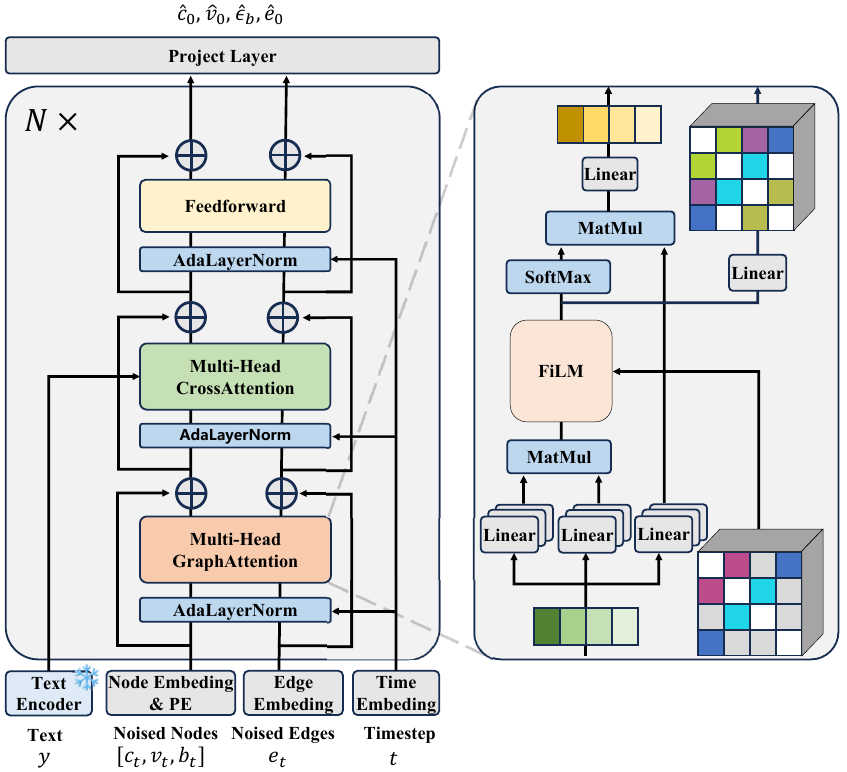}
\caption{Mixed Graph Diffusion Transformer. MG-DiT conditions on text and timesteps to jointly denoise continuous bounding box features along with discrete graph and fVQ-VAE features. The network consists of input and output processing layers, stacked with several Mixed Graph Transformer Blocks, as illustrated.}
\label{fig:network}
\vspace{-15pt}
\end{figure}

\noindent\textbf{Mixed Graph Diffusion.} 
Drawing inspiration from MiDiffusion~\cite{hu2024mixed}, we inject domain-specific noise into both continuous and discrete variables simultaneously during training. For a scene in the form of a graph, we have $\boldsymbol{b}=[\boldsymbol{s}, \boldsymbol{t}, \boldsymbol{r}]$ to represent the continuous attributes, i.e. object size, position, and orientation, and $\boldsymbol{z}=[\boldsymbol{c}, \boldsymbol{v}, \boldsymbol{e}]$ to denote the discrete attributes, i.e. object category, feature indices (fVQ-VAE indies), and the relationship category. Then the mixed forward diffusion process can be expressed as follows:
\begin{small}
\begin{equation}
q\left(\boldsymbol{b_t},\boldsymbol{z}_t|\boldsymbol{b}_{t-1},\boldsymbol{z}_{t-1}\right):=\tilde{q}(\boldsymbol{b}_t|\boldsymbol{b}_{t-1})\cdot\hat{q}(\boldsymbol{z}_t|\boldsymbol{z}_{t-1}).
\end{equation}
\end{small}%
Correspondingly, we can factor the posterior distribution of the denoising process as:
\begin{small}
\begin{equation}
\begin{aligned}
q\left(\boldsymbol{b}_{t-1},\boldsymbol{z}_{t-1}|\boldsymbol{b}_t,\boldsymbol{z}_t,\boldsymbol{b}_0,\boldsymbol{z}_0\right)=&\ \tilde{q}(b_{t-1}|b_t,b_0)\\
&\cdot\hat{q}(\boldsymbol{z}_{t-1}|\boldsymbol{z}_t,\boldsymbol{z}_0).
\end{aligned}
\end{equation}
\end{small}%
We introduce MG-DiT to fit this posterior distribution:
\begin{small}
\begin{equation}
\begin{aligned}
p_\theta\left(\boldsymbol{b}_{t-1},\boldsymbol{z}_{t-1}|\boldsymbol{b}_t,\boldsymbol{z}_t,\boldsymbol{y}\right):=&\ \tilde{p}_\theta(b_{t-1}|\boldsymbol{b}_t,\boldsymbol{z}_t,\boldsymbol{y})\\
&\cdot\hat{p}_\theta(\boldsymbol{z}_{t-1}|\boldsymbol{b}_t,z_t,\boldsymbol{y}),
\end{aligned}
\end{equation}
\end{small}%
where $\theta$ is the MG-DiT network parameter. The negative log-likelihood is employed as our loss function. Following DDPM~\cite{ho2020denoising}, we adopt the variational lower bound and omit the constant term $L_T$:
\begin{small}
\begin{equation}
\begin{aligned}
\mathcal{L}=&\ \mathcal{L}_{b}+\mathcal{L}_{z}\\
=&\ \mathbb{E}_{\boldsymbol{b}_{0},\boldsymbol{z}_{0},\boldsymbol{t}\in[1,T],\boldsymbol{\epsilon}_b,q(\boldsymbol{z}_{t}|\boldsymbol{z}_{0})}[\\
&\ \|\boldsymbol{\epsilon}_b-\boldsymbol{\epsilon}_\theta(\sqrt{\bar{\alpha}_t}\boldsymbol{b}_0+\sqrt{1-\bar{\alpha}_t}\boldsymbol{\epsilon}_b,t,\boldsymbol{y})\|^2\\
&\ +D_{KL}(\tilde{q}(\boldsymbol{z}_{t-1}|\boldsymbol{z}_{t},\boldsymbol{z}_{0})||\tilde{p}_{\theta}(\boldsymbol{z}_{t-1}|\boldsymbol{z}_{t}, \boldsymbol{y}))],
\end{aligned}
\end{equation}
\end{small}%
where $t$ is sampled from a uniform distribution $\mathcal{U}(1,T)$ and $\epsilon$ is sampled from a standard normal distribution $\mathcal{N}(0,\mathbf{I})$.
\noindent\textbf{Network Architecture.}
As depicted in Figure \ref{fig:network}, in each timestep of the denoising process, we embed and concatenate the noisy attributes $(c,v,s,t,r)$ as node features, incorporating sinusoidal positional encodings. Simultaneously, the noisy attribute $e$ is embedded as edge features. These nodes and edges are fed into a DiT network \cite{peebles2023scalable} to fit the posterior distribution, in which we integrate the cross-attention modules to handle text input. Following \cite{dwivedi2020generalization,vignac2022digress,lin2024instructscene}, we employ FiLM~\cite{perez2018film} as the interaction mechanism between nodes and edges, which can be expressed by the following function:
\begin{small}
\begin{equation}
\mathrm{FilM}(\boldsymbol{sim},\boldsymbol{e})=\gamma(\boldsymbol{e})\left(\frac{\boldsymbol{sim}-\mu(\boldsymbol{sim})}{\sigma(\boldsymbol{sim})}\right)+\beta(\boldsymbol{e}), 
\end{equation}
\end{small}%
where $\gamma(\cdot)$ and $\beta(\cdot)$ represent the mapping from the edge matrix to the new scale and bias. The network's output consists of the predicted noise $\hat{\epsilon}_{s}, \hat{\epsilon}_{r}, \hat{\epsilon}_{r}$ for the bounding box, as well as the predicted values of $\hat{c}_{0}$, $\hat{v}_{0}$, and $\hat{e}_{0}$ at timestep 0, which are used to calculate the posterior probability distribution during the denoising process.

\subsection{Zero-Shot Applications}
\label{application}
Constrained sampling has been employed in prior works \cite{paschalidou2021atiss, tang2024diffuscene, lin2024instructscene} for tasks such as scene re-arrangement and completion to maintain some fixed elements. For instance, in re-arrangement tasks, the categories and sizes of objects are fixed while their positions are perturbed and denoised. In completion tasks, all attributes of existing objects are fixed while the rest of objects are generated. We extend this approach by fixing the graph defined by the graph designer or users, while generating the remaining attributes. This enables the use of a single model for both text-to-scene and graph-to-scene generation, without the need for additional training. Besides, we can apply MG-DiT to other zero-shot downstream applications. The variables for noising and denoising process encompass $c$, $v$, $b$ and $e$, where $b=[s,t,o]$. At each step of the sampling process, we fix the corresponding variables to support different applications, and the settings of different applications are presented in Table \ref{tab1}. For simplicity, we use the same notation as in Section \ref{problem_formulation} but omit the subscripts and superscripts.
\begin{table}[t]
\centering
\scalebox{0.86}{
\begin{tabular}{cp{0.8cm}<{\centering}p{0.8cm}<{\centering}p{0.8cm}<{\centering}p{0.8cm}<{\centering}p{0.8cm}<{\centering}}
\toprule[1.3pt]
\diagbox{\textbf{App.}}{\textbf{Var.}}&$c$&$e$&$s$&$[t,o]$&$v$ \\
\midrule[1pt]
\textbf{Text-to-scene}&gen&gen&gen&gen&gen\\
\textbf{Graph-to-scene}&f/p&f/p&gen&gen&gen\\
\textbf{Re-arrangement}&fixed&gen&fixed&gen&fixed\\
\textbf{Completion}&partial&partial&partial&partial&partial\\
\textbf{Stylization}&fixed&fixed&fixed&fixed&gen\\
\bottomrule[1.3pt]
\end{tabular}
}
\caption{Feature fixation solutions for different applications. `gen' represents normal denoising. `fixed' indicates that the features of all objects in a scene remain unchanged during the denoising process, while `partial' signifies that only partial existing features are fixed. `f/p' means we can either fix all attributes for a full condition, or only fix partial existing attributes for a partial condition.}
\label{tab1}
\vspace{-15pt}
\end{table}
\section{Experiments}

\noindent \textbf{Datasets.}
Following previous works, we conduct experiments on 3D-FRONT~\cite{fu20213dfr} dataset, an indoor scene dataset composed of 6,813 houses with 14,629 rooms.
During the object retrieval phase of our indoor scene synthesis, we select the closest furniture models from 3D-FUTURE~\cite{fu20213dfu} based on object categories and fVQ-VAE features.
We conduct experiments on three types of scenes: 4,041 bedrooms, 900 dining rooms, and 813 living rooms. For further details on data preprocessing, please refer to the Supplementary.

\noindent \textbf{Baselines.}
To validate the effectiveness of our approach, we evaluate its performance from \textit{three} perspectives: text-to-scene generation, zero-shot graph-to-scene generation, and other applications. 
We compare our method with the state-of-the-art approaches including: 
\textbf{1)} ATISS \cite{paschalidou2021atiss}, a transformer-based autoregressive model to which we added a text-based cross-attention layer to enable text-to-scene generation; 
\textbf{2)} DiffuScene \cite{tang2024diffuscene}, a diffusion model for indoor scene synthesis with the U-Net architecture; 
\textbf{3)} InstructScene \cite{lin2024instructscene}, a dual-model approach that first predicts the graph using a text-to-graph model and then generates the scene using a graph-to-scene model. 

\noindent \textbf{Metrics.} 
We focus on two key aspects for a comprehensive evaluation of our approach.
\textbf{1) For the quality of the generated scenes}, we primarily assess the following metrics: 
Fréchet Inception Distance (FID) (\cite{heusel2017gans}), $\text{FID}^{\text{CLIP}}$ \cite{kynkaanniemi2022role} (which uses CLIP features \cite{radford2021learning} to compute $\text{FID}$ scores), Kernel Inception Distance (KID) \cite{binkowski2018demystifying}, and Scene Classification Accuracy (SCA) \cite{paschalidou2021atiss}. SCA involves training a discriminator model to distinguish between real and generated data, assessing the ability of the generated data to fool the discriminator. The closer the model's accuracy on generated data is to 50\%, the more realistic the generated results are perceived.
\textbf{2) For the controllability of the models}, we adopt and extend the iRecall metric introduced in InstructScene \cite{lin2024instructscene}. For text-to-scene generation, iRecall measures the probability that each sentence in the input text occurs in the generated results. For graph-to-scene generation, iRecall reflects the probability that each relationship in the graph occurs in the generated scene.

%
\subsection{Evaluation for Graph Designer} 
To validate the effectiveness of Graph Designer, we compared the graph extraction results of one-shot CoT prompts and one-shot prompts without CoT across a collection of images and text, including the photos and sketches of different views and room types, manually created top-view diagrams, and a set of textual scene descriptions. Both of the two prompt templates comprise a formatted example to ensure the consistency of the results. The detailed prompt templates can be found in Supplementary. Additionally, during testing, we input either images or text into the VLMs to evaluate the performance on each specific modality. We assess the extraction performance from two aspects: the ratio of correctly identified objects w.r.t. the total number of objects named \textbf{Object iRecall} and the proportion of correctly recognized relationships named \textbf{Rel Acc}. 

\begin{table}[t]
\centering
\scalebox{0.8}{
\begin{tabular}{cccc}
\toprule[1.3pt]
\multicolumn{2}{c}{\textbf{Graph Designer}}&$\uparrow\textbf{Object iRecall}_{\%}$&$\uparrow\textbf{Rel Acc}_{\%}$\\
\midrule[1pt]
\multirow{4}{*}{\parbox{1.8cm}{with CoT vs.\\without CoT}}
&\textbf{Image}&85.23/72.07&77.56/34.65\cr
&\textbf{Sketch}&64.97/58.25&72.15/31.97\cr
&\textbf{Diagram}&91.22/88.13&74.63/44.32\cr
&\textbf{Text}&98.56/95.56&89.45/85.60\\
\bottomrule[1.3pt]
\end{tabular}
}
\caption{Quantitative comparisons on graph extraction performance between one-shot CoT prompts (with CoT) and one-shot prompts (without CoT). For each pair of data, number before and after the slash is with CoT and without CoT, respectively.}
\vspace{-15pt}
\label{tab2}
\end{table}

Table \ref{tab2} shows the quantitative evaluations of one-shot CoT prompts and one-shot prompts without CoT. As demonstrated, one-shot CoT prompts can significantly enhance the graph extraction performance of the VLMs across various modalities, particularly in terms of relationship accuracy. This is because one-shot prompts are prone to producing contradictory or erroneous relationships. In contrast, one-shot CoT prompts can effectively guide the VLM to hierarchically navigate the entire scene, precisely establishing accurate relationships throughout. We also present a comparison of the graph extraction performance on a specific image-text pair example between one-shot CoT prompts and one-shot prompts, the results are provided in Supplementary due to space constraints.
\begin{figure*}
\centering
\includegraphics[width=1\linewidth]{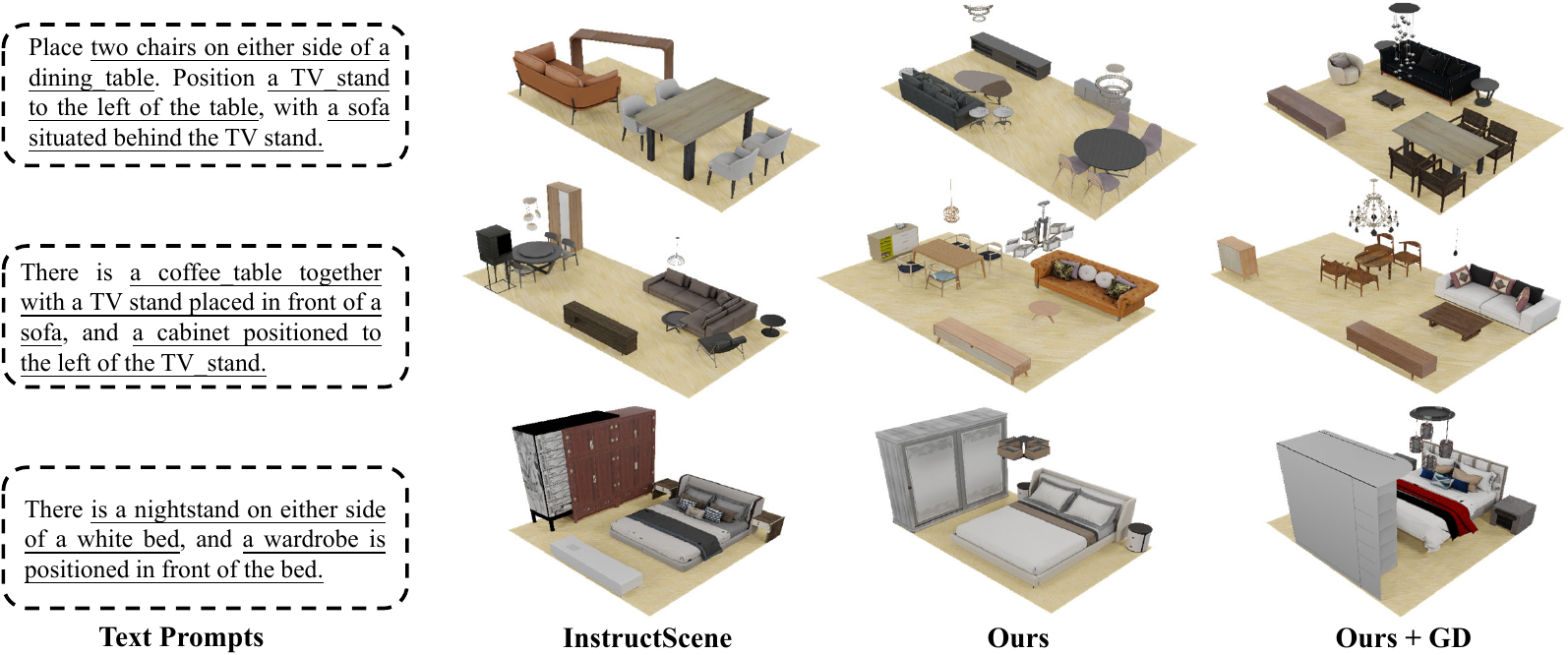}
\vspace*{-5pt}
\caption{Qualitative comparisons on text-to-scene generation.}
\label{fig:qualitative}
\end{figure*}
\subsection{Text-to-Scene Generation}
We compare with three existing methods to demonstrate the effectiveness of our method. Additionally, we provide two configurations with our approach: the first directly employs MG-DiT for text-to-scene generation (named \textbf{Ours}); the second uses Graph Designer to preprocess the textual scene description, followed by the MG-DiT conditioned on the extracted partial graphs to generate the scenes (named \textbf{Ours+GD}). 
To ensure a fair comparison, we keep the floor texture and background consistent among all the results. 
\begin{table}[t]
\centering
\resizebox{1\columnwidth}{!}{
\setlength{\tabcolsep}{1.8pt}
\begin{tabular}{ccccccc}
\toprule[1.3pt]
\multicolumn{2}{c}{
\textbf{Text-to-scene}}&\textbf{FID}&$\textbf{FID}^\textbf{CLIP}$&$\textbf{KID}_{\times{1e^{-3}}}$&$\textbf{SCA}_{\%}$&$\textbf{iRecall}_{\%}$\\
\midrule[1pt]
\multirow{5}{*}{\textbf{Bedroom}}
&\textbf{ATISS}&122.37&8.23&0.74&58.77&44.06\cr 
&\textbf{DiffuScene}&129.34&9.66&0.81&59.13&53.32\cr
&\textbf{InstructScene}&114.86&6.52&0.68&56.37&72.71\cr
&\textbf{Ours}&111.21&6.43&0.35&54.94&73.69\cr
&\textbf{Ours+GD}&\textbf{108}&\textbf{6.07}&\textbf{0.21}&\textbf{53.16}&\textbf{81.40}\\
\midrule
\multirow{5}{*}{\textbf{Livingroom}}
&\textbf{ATISS}&120.10&7.43&16.44&57.02&36.31\cr 
&\textbf{DiffuScene}&135.93&10.71&29.07&58.98&39.58\cr
&\textbf{InstructScene}&111.52&5.91&8.65&55.32&57.21\cr
&\textbf{Ours}&110.55&5.83&7.95&55.24&58.16\cr
&\textbf{Ours+GD}&\textbf{108.22}&\textbf{5.23}&\textbf{3.87}&\textbf{54.05}&\textbf{71.81}\\
\midrule
\multirow{5}{*}{\textbf{Diningroom}}
&\textbf{ATISS}&134.75&9.82&24.25&59.83&32.56\cr 
&\textbf{DiffuScene}&142.37&11.76&28.36&60.88&37.17\cr
&\textbf{InstructScene}&129.13&8.24&15.27&58.93&61.47\cr
&\textbf{Ours}&127.28&8.01&14.83&56.82&63.39\cr
&\textbf{Ours+GD}&\textbf{125.32}&\textbf{7.14}&\textbf{10.68}&\textbf{55.59}&\textbf{75.01}\\
\bottomrule[1.3pt]
\end{tabular}
}
\caption{Quantitative comparisons on text-to-scene generation. Lower FID, $\text{FID}^\text{CLIP}$, KID and higher iRecall indicate better synthesis quality. For SCA, a score closer to 50\% is better.}
\vspace{-10pt}
\label{tab3}
\end{table}

Table \ref{tab3} presents a quantitative comparison of our method against other approaches across various metrics. As shown in the table, our method exhibits excellent text-to-scene generation performance and surpasses other approaches in various metrics. This reveals that taking graph information into account significantly improves the feature extraction ability of our method during the generation process, making it outperform the autoregressive model ATISS and the diffusion model DiffuScene by a considerable margin. The 1D convolution of DiffuScene makes it challenging to capture global features, while ATISS struggles to integrate global text descriptions, leading to reduced controllability. Compared to InstructScene, since our MG-DiT model employs a combination of discrete graph and continuous bounding box variables, it has to predict the graph from discrete noise, which contains [MASK] tokens or other noise states, while concurrently generating the bounding boxes. This forces the model to more effectively learn global scene features and grasp the relationships between objects during graph prediction, thereby improving the final generation quality and controllability. By directly extracting graph from text using Graph Designer as a preprocessing step, our model can achieve superior generation quality and controllability. 

In Figure \ref{fig:qualitative}, we present a qualitative comparison with several baselines. It is evident that our MG-DiT algorithm produces higher-quality layouts compared to InstructScene. Moreover, preprocessing the text using the Graph Designer enhances controllability, yielding outputs that align more closely with the input text. Additional results are provided in the Supplementary Materials due to space constraints.


\subsection{Graph-to-Scene Generation}
Regarding graph-to-scene generation, we assess both the quality of the output scenes and the consistency between the generated scenes and the provided graphs.  Since ATISS and DiffuScene do not involve graph priors, we only compared the graph-conditioned results of MG-DiT with InstructScene.
For InstructScene, we employ constrained sampling in the text-to-graph model to fix the graph (object categories and edges) and generate fVQ-VAE indices, then use the graph-to-scene model to produce the final results. Table \ref{tab4} shows the quantitative results, indicating that our hybrid graph diffusion architecture achieves improved performance in both generation quality and controllability. More qualitative results are presented in Supplementary.
\begin{table}[t]
\centering
\resizebox{1\columnwidth}{!}{
\setlength{\tabcolsep}{1.8pt}
\begin{tabular}{ccccccc}
\toprule[1.3pt]
\multicolumn{2}{c}{
\textbf{Graph-to-scene}}&\textbf{FID}&$\textbf{FID}^\textbf{CLIP}$&$\textbf{KID}_{\times{1e^{-3}}}$&$\textbf{SCA}_{\%}$&$\textbf{iRecall}_{\%}$\\
\midrule[1pt]
\multirow{2}{*}{\textbf{Bedroom}}
&\textbf{InstructScene}&101.86&5.66&0.13&53.68&88.84\cr
&\textbf{Ours}&\textbf{98.31}&\textbf{5.58}&\textbf{0.12}&\textbf{52.34}&\textbf{89.37}\\
\midrule
\multirow{2}{*}{\textbf{Livingroom}}
&\textbf{InstructScene}&108.45&5.43&0.33&54.51&83.14\cr
&\textbf{Ours}&\textbf{107.62}&\textbf{5.25}&\textbf{0.29}&\textbf{53.77}&\textbf{83.35}\\
\midrule
\multirow{2}{*}{\textbf{Diningroom}}
&\textbf{InstructScene}&123.82&8.21&14.18&56.23&84.16\cr
&\textbf{Ours}&\textbf{122.57}&\textbf{7.49}&\textbf{13.11}&\textbf{55.70}&\textbf{84.47}\\
\bottomrule[1.3pt]
\end{tabular}
}
\caption{Quantitative comparisons on graph-to-scene generation}
\vspace{-8pt}
\label{tab4}
\end{table}

\subsection{More Applications}
Table \ref{tab5} presents comparisons with existing methods across various applications, including re-arrangement, completion, stylization and unconditioned generation without any text or graph condition. To evaluate scene stylization, we use metric $\Delta$ in \cite{lin2024instructscene} which measures the cosine similarity between the CLIP features of the target style text and the top-down view images of the generated scenes. Since ATISS and DiffuScene lack features indicating color and texture, it is challenging to conduct stylization experiments. Additionally, as an autoregressive model, ATISS is unsuitable for re-arrangement tasks. Therefore, we have excluded these experimental results. Our experimental findings clearly demonstrate that our method outperforms other baseline approaches across re-arrangement, completion, and unconditional generation tasks and achieves comparable results to InstructScene in the stylization task. More qualitative results are presented in Supplementary.

\begin{table}[t]
\centering
\resizebox{1\columnwidth}{!}{
\setlength{\tabcolsep}{1.3pt}
\begin{tabular}{cccccccc}
\toprule[1.3pt]
\multicolumn{2}{c}{
\multirow{2}{*}{\textbf{Applicaitons}}}&\multicolumn{2}{c}{\textbf{Re-arrangement}}&\multicolumn{2}{c}{\textbf{Completion}}&\textbf{Stylization}&\textbf{Uncond.}\cr
\multicolumn{2}{c}{}&\textbf{FID}&$\textbf{iRecall}_{\%}$&\textbf{FID}&$\textbf{iRecall}_{\%}$&$\Delta_{\times1e-3}$&$\textbf{FID}$\\
\midrule[1pt]
\multirow{4}{*}{\textbf{Bedroom}}
&\textbf{ATISS}&-&-&95.51&43.77&-&126.80\cr 
&\textbf{DiffuScene}&119.46&55.37&100.72&52.15&-&138.24\cr
&\textbf{InstructScene}&106.80&76.24&82.94&67.47&6.55&124.77\cr
&\textbf{Ours}&\textbf{105.71}&\textbf{77.49}&\textbf{81.83}&\textbf{69.25}&\textbf{6.79}&\textbf{123.42}\\
\midrule
\multirow{4}{*}{\textbf{Livingroom}}
&\textbf{ATISS}&-&-&99.25&28.65&-&124.55\cr 
&\textbf{DiffuScene}&124.65&32.33&103.17&25.08&-&139.63\cr
&\textbf{InstructScene}&107.78&53.40&92.38&48.53&0.29&117.71\cr
&\textbf{Ours}&\textbf{106.33}&\textbf{54.82}&\textbf{91.01}&\textbf{50.57}&\textbf{0.31}&\textbf{116.87}\\
\midrule
\multirow{4}{*}{\textbf{Diningroom}}
&\textbf{ATISS}&-&-&117.20&35.70&-&139.26\cr 
&\textbf{DiffuScene}&136.37&34.62&121.53&33.97&-&148.54\cr
&\textbf{InstructScene}&125.96&62.45&107.44&59.21&1.83&136.40\cr
&\textbf{Ours}&\textbf{124.69}&\textbf{63.02}&\textbf{105.71}&\textbf{60.67}&\textbf{1.87}&\textbf{132.22}\\
\bottomrule[1.3pt]
\end{tabular}
}
\caption{Quantitative comparisons on various applications.}
\label{tab5}
\vspace{-8pt}
\end{table}

\vspace{-5pt}
\section{Conclusion}
\vspace{-3pt}
We introduce FreeScene, a novel framework designed to promote both user interactivity and the quality of indoor scene synthesis through the integration of a VLM-based Graph Designer and the MG-DiT model. Experimental results demonstrate our FreeScene can not only enhance the performance of various tasks, but also provides a more user-friendly control mechanism for scene synthesis.

\noindent \textbf{Limitaions \& Future Work.}
FreeScene also has several limitations. First, the Graph Designer is limited by the capabilities of VLMs, which may struggle to capture all the graph priors, especially in scenes with a large number of objects.
This limitation could be mitigated with future advancements in VLM technology. Second, the MG-DiT cannot precisely control the exact position and orientation of objects, which may result in furniture arrangements that are impractical for human use. 
Adding additional constraints, like affordability, to diffusion models may offer a promising solution to this issue. Last but not least, the scale and quality of available indoor scene datasets are still insufficient, which may lead to overfitting and suboptimal performance. A potential direction for future work to address this data limitation is to combine the general knowledge embedded in large models with real samples from databases, leveraging techniques such as Retrieval-Augmented Generation (RAG) for interactive room design.

\section*{Acknowledgements}
This research was supported in part by the National Natural Science Foundation of China (No. 62202199, 62302269), the Excellent Young Scientists Fund Program (Overseas) of Shandong Province (No. 2023HWYQ-034) and the Fundamental Research Funds for the Central Universities.

{
    \small
    \bibliographystyle{ieeenat_fullname}
    \bibliography{main}
}

\clearpage
\setcounter{page}{1}
\maketitlesupplementary


\setcounter{section}{0} 
\renewcommand{\thesection}{\Alph{section}}
\setcounter{figure}{0} 

\section{Implementation Details}
\label{sec:impledetail}
\subsection{Data Processing}
\begin{figure}[H]
\centering
\includegraphics[width=1\linewidth]{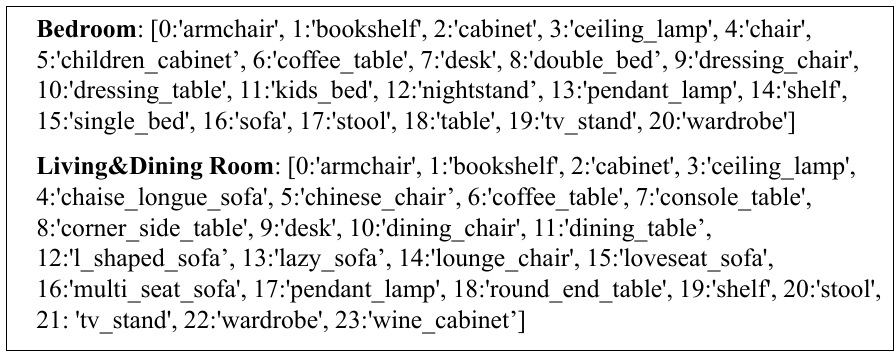}
\caption{Category lists for different room types.}
\label{fig:obrange}
\end{figure}

\noindent \textbf{Edge Matrix Construction.} In this paper, we define ten types of spatial relationships and a `None' type, similar to InstructScene, including: `Left of', `Right of', `In front of', `Behind', `Closely left of', `Closely right of', `Closely in front of', `Closely behind', `Above', `Below' and `None'. Additionally, based on the dataset, we defined different category list tailored to different room types as shown in Figure \ref{fig:obrange}. Upon receiving the triplet list output from the Graph Designer, we construct a specialized symmetric matrix where each pair of symmetric positions stores opposite spatial relationships, such as `Left of' and `Right of'.
\noindent \textbf{Text Prompt Generation.} We followed the methodology of InstructScene to generate text prompts for each scene, which include spatial relations extracted from the 3DFront dataset and object captions obtained using BLIP. Finally, 1$\sim$2 of these relationships are randomly selected and refined using ChatGPT.

\subsection{Model Details}

\noindent \textbf{Training Strategies.}
We employ a network containing five Transformer blocks with 8-head 512-dimensional attention and a dropout rate of 0.1. 
In order to improve the robustness of the training process, we employ a probabilistic approach to selectively fix certain variables. Specifically, we assign a 0.2 probability to nullify the text input (for classifier-free guidance), a 0.2 probability to preserve the input graph (steering the model to predict noise that aligns with the graph), and a 0.1 probability to generate only the edges while keeping the other features fixed. The remaining 0.5 probability is dedicated to denoising all the variables. The models are trained using the AdamW optimizer for 500,000 iterations with a batch size of 128, a learning rate of 1e-4, and a weight decay of 0.02. 
All experiments are conducted on a single NVIDIA A40 48GB GPU.

\subsection{Prompt Templates}
\begin{figure*}
\centering
\includegraphics[width=1\linewidth]{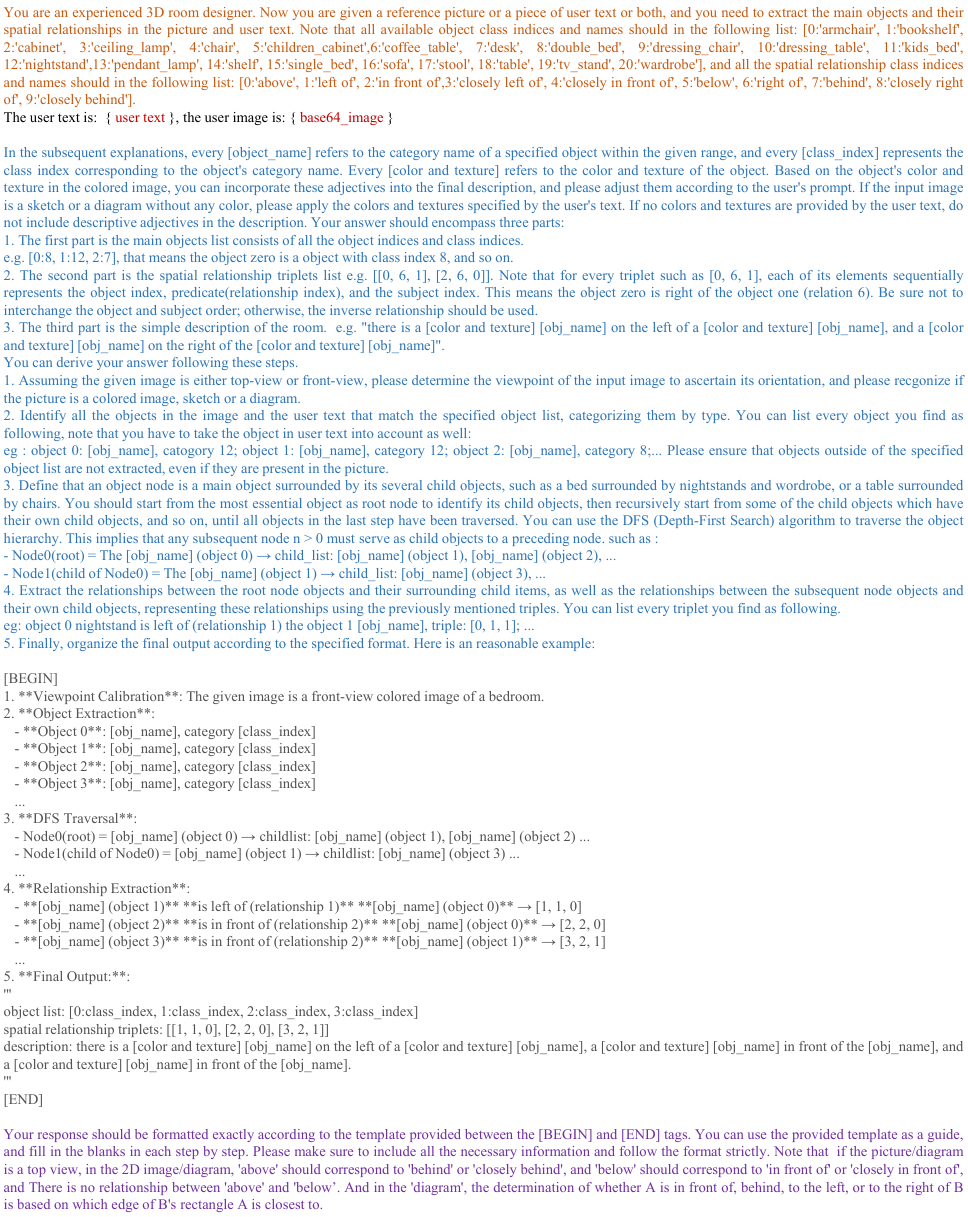}
\caption{One-shot CoT prompt template for bedroom.}
\label{fig:cot_prompt}
\end{figure*}

In Figure \ref{fig:cot_prompt}, we provide the one-shot Chain-of-Thought (CoT) prompt utilized by the Graph Designer, illustrated with the example of a bedroom. For room types such as dining room and living room, the prompt template can be effortlessly adapted by substituting the category list with the corresponding range.

Given the inherent randomness and variability in GPT-4o’s text and image comprehension, its outputs may occasionally be unprocessable or of suboptimal quality. To mitigate this, we employ a validation and retry mechanism, ensuring that within a few retries, a usable result can be obtained.

\section{Additional Results}
\subsection{Evaluation on One-Shot CoT Prompts}
\begin{figure}[H]
\centering
\includegraphics[width=1\linewidth]{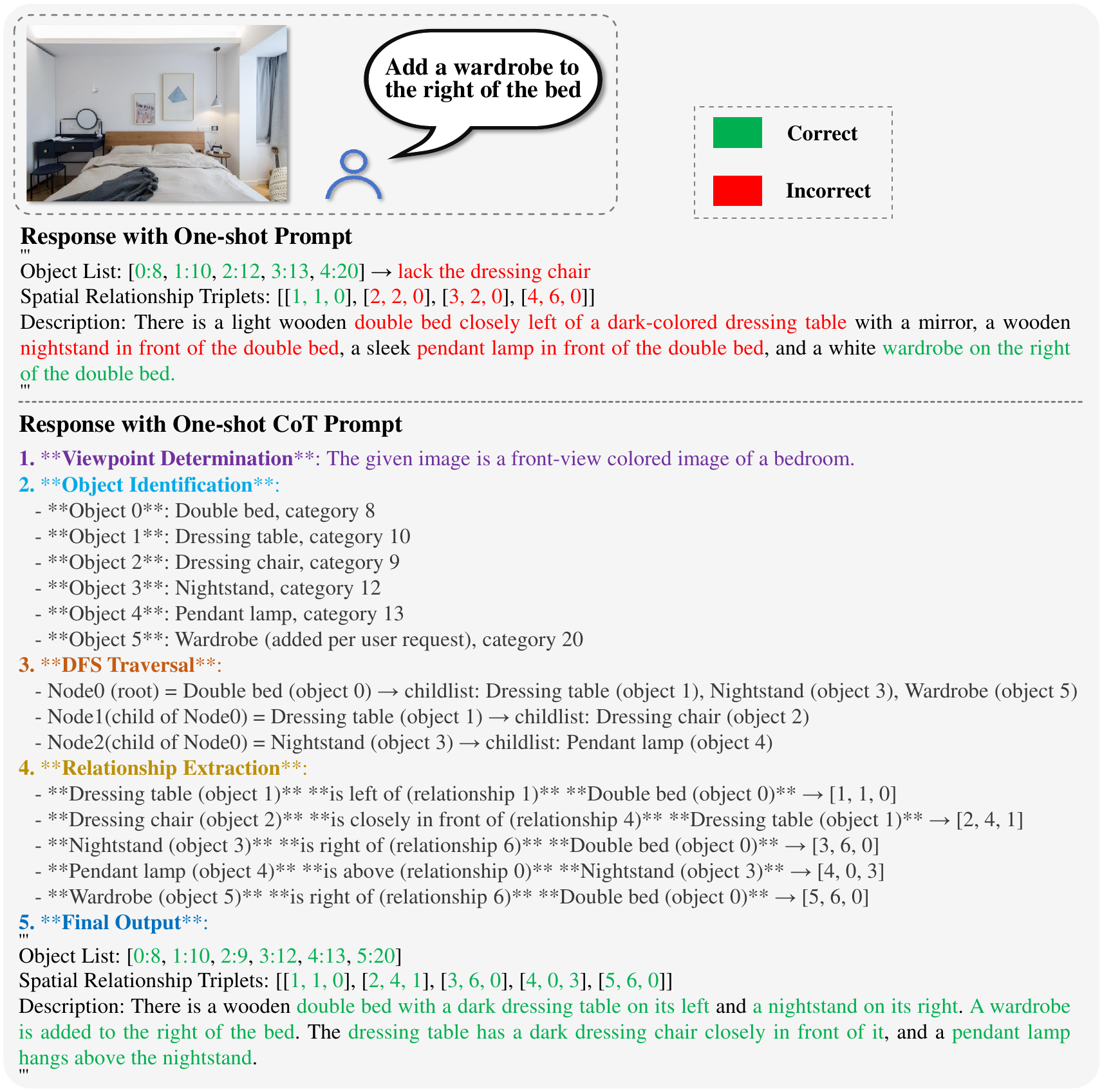}
\caption{A comparison of the results between one-shot CoT prompts and one-shot prompts on a specific example.}
\label{fig:gd_eval}
\end{figure}

In Figure \ref{fig:gd_eval}, we also present a comparison of the graph extraction results on an image-text pair using one-shot CoT prompts and one-shot prompts. It is clearly demonstrated that one-shot CoT prompts significantly enhance the performance of the Graph Designer through stepwise guidance, particularly in the accuracy of spatial relationships, making the graph extraction process more reliable.
\subsection{More Qualitative Results}
\begin{figure*}
\centering
\includegraphics[width=0.95\linewidth]{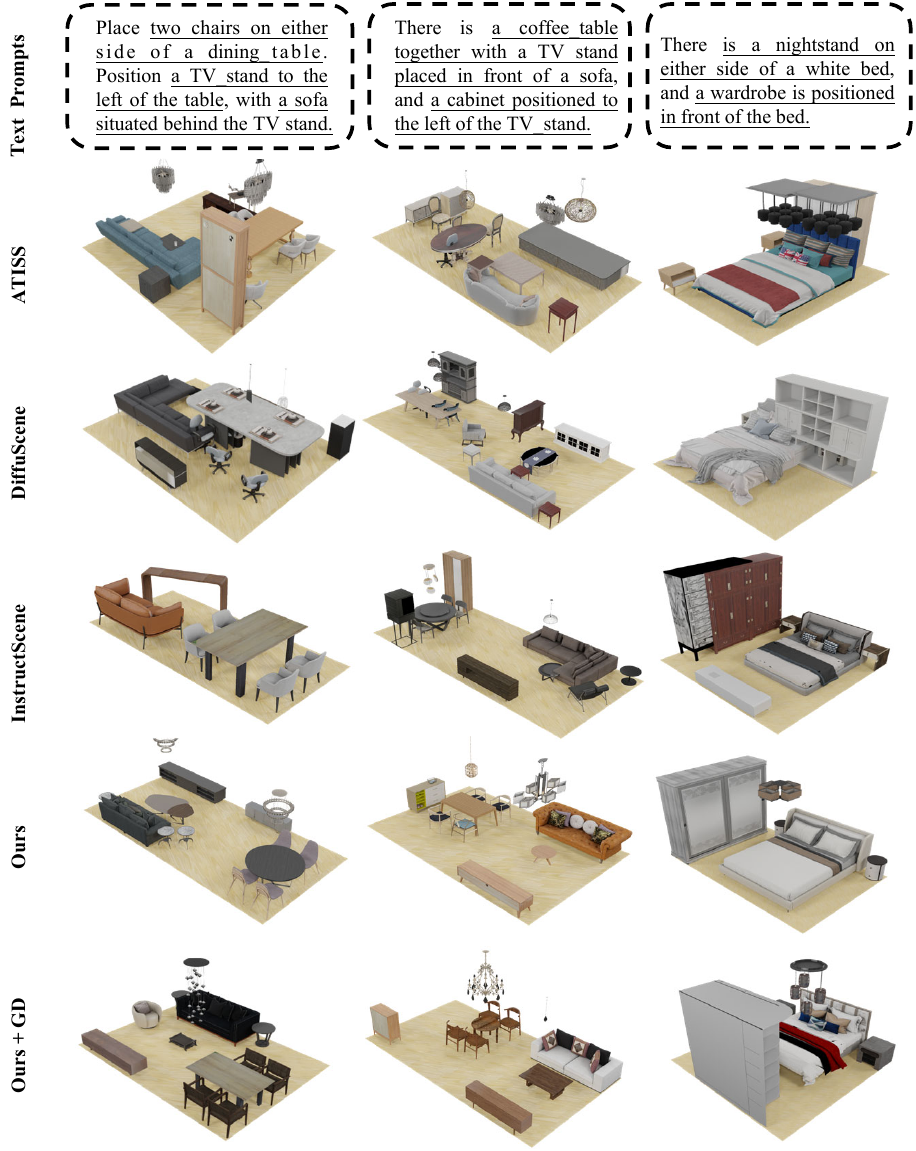}
\caption{Qualitative comparisons on text-to-scene generation.}
\label{fig:t2s_qualitative}
\end{figure*}

\begin{figure*}
\centering
\includegraphics[width=0.95\linewidth]{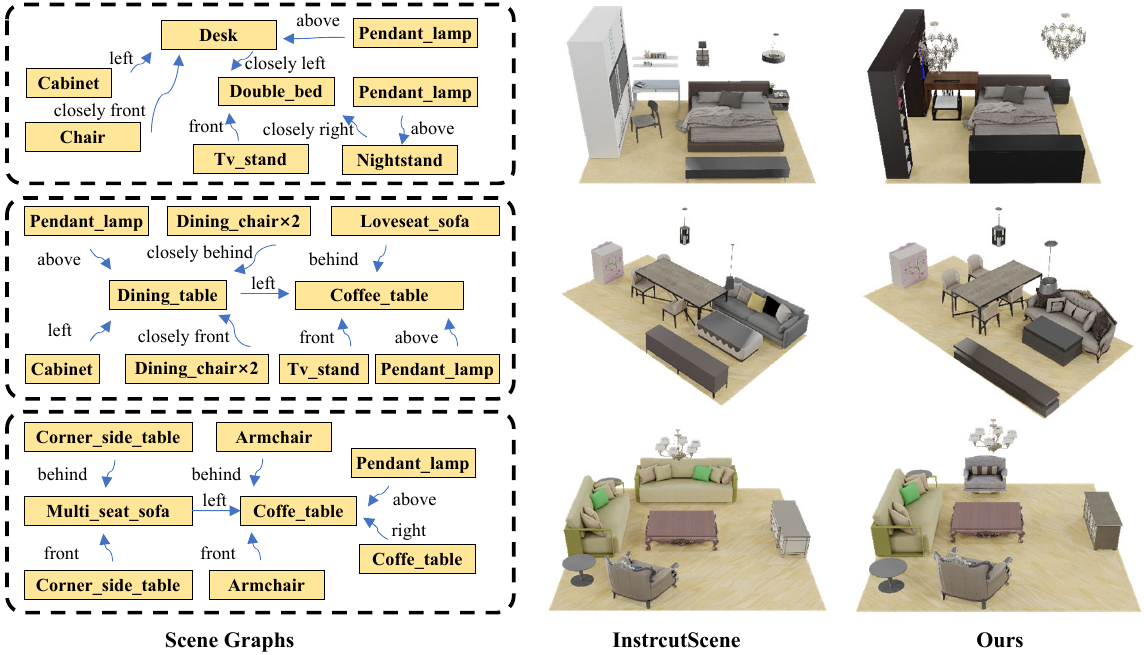}
\caption{Qualitative comparisons on graph-to-scene generation.}
\label{fig:g2s_qualitative}
\end{figure*}

\begin{figure*}
\centering
\includegraphics[width=0.95\linewidth]{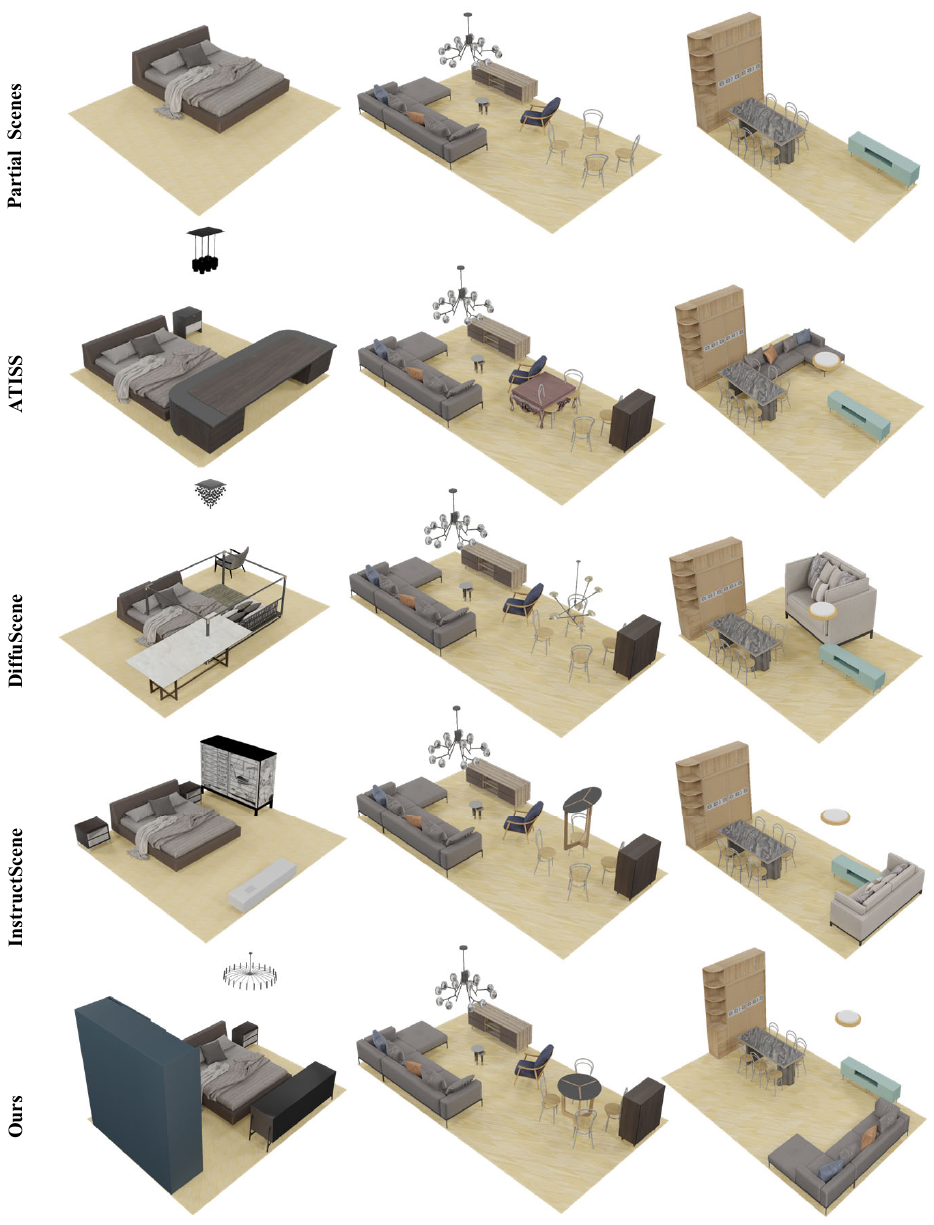}
\caption{Qualitative comparisons on completion. Note that, in this task, the main goal is to complete the scene by considering the given objects in the input partial scene and the floor size is automatically adapted to the boundary of the updated scene.}
\label{fig:com_qualitative}
\end{figure*}

In this section, we present additional qualitative results for text-to-scene, graph-to-scene, completion, and stylization tasks, which are showcased in Figures \ref{fig:t2s_qualitative}, \ref{fig:g2s_qualitative}, \ref{fig:com_qualitative}, and \ref{fig:st_qualitative}, respectively. All qualitative results are rendered using Blender's Python API.

\subsection{User Study}
To further evaluate these methods, we conducted a user study comparing InstructScene, MG-DiT, and full version of FreeScene.
As shown in Table \ref{tab_usr}, FreeScene demonstrates clear advantages over other methods
\begin{table}[t]
\centering

\small
\begin{tabular}{cccc}
\toprule
\textbf{Rank}\textbackslash\textbf{Method}&\textbf{FreeScene}&\textbf{MG-DiT}&\textbf{InstructScene}\cr
$\textbf{Top1}_{\%}$&57.10&24.52&18.39\cr
$\textbf{Top2}_{\%}$&87.74&63.23&49.03\\
\bottomrule
\end{tabular}
\vspace{-0.2cm}
\caption{The user study involving 31 participants to evaluate 10 sets of scenes generated by three methods and focused on two key aspects: text-scene consistency and scene plausibility. For each result, we provided a top-down view and a randomly selected perspective. Participants were asked to rank the outputs from the three methods based on the given criteria, from best to worst. We subsequently analyzed and calculated the probability of each method being ranked as Top1 and Top2 across all evaluations.}
\vspace{-8pt}
\label{tab_usr}
\end{table}

\twocolumn[{%
\renewcommand\twocolumn[1][]{#1}%
\begin{center}
\centering
\includegraphics [width=0.95\linewidth]{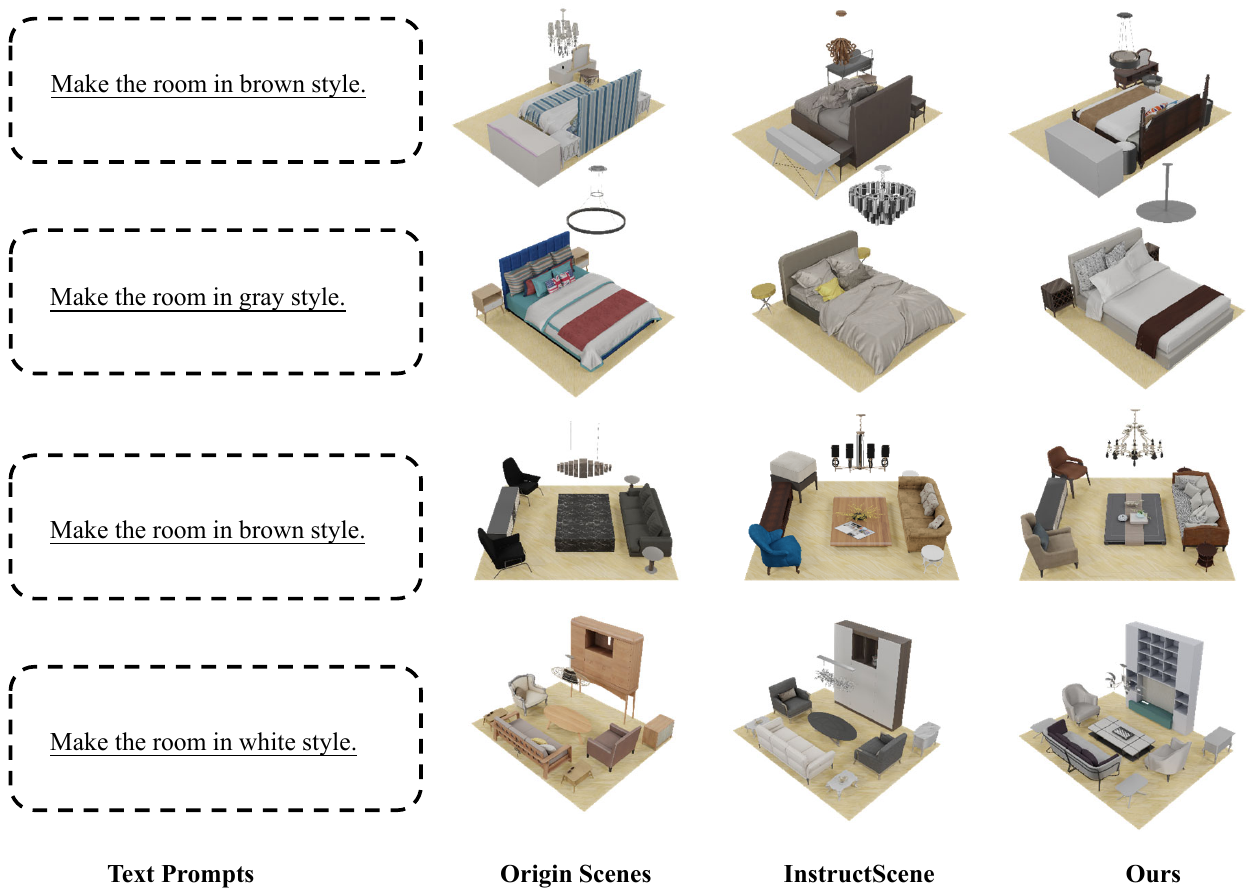}
\captionof{figure}{Qualitative comparisons on stylization.}
\label{fig:st_qualitative}
\end{center}
}]



\end{document}